%% file: main.tex
\documentclass[10pt,twocolumn,letterpaper]{article}

\usepackage[pagenumbers]{cvpr} % To force page numbers, e.g. for an arXiv version

\input{preamble}
\usepackage{xcolor}    
\definecolor{cvprblue}{rgb}{0.21,0.49,0.74}
\usepackage[pagebackref,breaklinks,colorlinks,allcolors=cvprblue]{hyperref}     % colors
\usepackage{adjustbox}
\usepackage{wrapfig}
\usepackage{multirow}
\usepackage{amsmath}
\usepackage{mathtools}
\usepackage{enumitem}
\usepackage{makecell}
\usepackage{colortbl}
\usepackage{pifont}
\usepackage{subcaption}
\usepackage{amssymb}
\usepackage{tabularx}
\usepackage{stfloats}
\usepackage[capitalize]{cleveref}
\usepackage{capt-of}

\makeatletter
\def\smallunderbrace#1#2{%
  \mathop{\vtop{\m@th\ialign{##\crcr
    $\hfil\displaystyle#1\hfil$\crcr
    \noalign{\kern3\p@\nointerlineskip}%
    {\tiny\upbracefill}\crcr\noalign{\kern3\p@}}}}\limits_{#2}}
\makeatother

\definecolor{mask_orange}{HTML}{FF6402}
\definecolor{lightblue}{RGB}{180,200,220}

\title{Long-RVOS: A Comprehensive Benchmark for Long-term Referring Video Object Segmentation}\vspace{-1em}

\author{%
  Tianming Liang$^{1}$, \quad 
  Haichao Jiang$^{1}$, \quad
  Yuting Yang$^{1}$, \quad
  Chaolei Tan$^{1}$, \quad
  Shuai Li$^{2}$,
  \\
  Wei-Shi Zheng$^{1}$, \quad
  Jian-Fang Hu$^{1\ast}$
  \\
  $^{1}$Sun Yat-sen University, \quad $^{2}$Shandong University.\\
  \texttt{liangtm@mail2.sysu.edu.cn, hujf5@mail.sysu.edu.cn}\\\vspace{-1em}\\
  \textbf{Project Page:} \url{https://isee-laboratory.github.io/Long-RVOS}
}

\begin{document}
\maketitle
\input{sec/0_abstract}    
\input{sec/1_intro}
\input{sec/2_related_work}
\input{sec/3_dataset.tex}

\input{sec/4_method.tex}
\input{sec/5_experiment.tex}

{
    \small
    \bibliographystyle{ieeenat_fullname}
    \bibliography{main}
}

% WARNING: do not forget to delete the supplementary pages from your submission 
\input{sec/X_suppl}

\end{document}

%% file: sec/0_abstract.tex
\begin{abstract}
Referring video object segmentation (RVOS) aims to identify, track and segment the objects in a video based on language descriptions, which has received great attention in recent years. 
However, existing datasets remain focus on short video clips within several seconds, with salient objects visible in most frames.
To advance the task towards more practical scenarios, we introduce \textbf{Long-RVOS}, a large-scale benchmark for long-term referring video object segmentation. 
Long-RVOS contains 2,000+ videos of an average duration exceeding 60 seconds, covering a variety of objects that undergo occlusion, disappearance-reappearance and shot changing.
The objects are manually annotated with three different types of descriptions to individually evaluate the understanding of static attributes, motion patterns and spatiotemporal relationships.
Moreover, unlike previous benchmarks that rely solely on the per-frame spatial evaluation, we introduce two new metrics to assess the temporal and spatiotemporal consistency.
We benchmark 7 state-of-the-art methods on Long-RVOS. The results show that current approaches struggle severely with the long-video challenges.
To address this, we further propose ReferMo, a promising baseline method that integrates motion information to expand the temporal receptive field, and employs a local-to-global architecture to capture both short-term dynamics and long-term dependencies.
Despite simplicity, ReferMo achieves significant improvements over current methods in long-term scenarios. 
We hope that Long-RVOS and our baseline can drive future RVOS research towards tackling more realistic and long videos. 
% Our dataset and code will be released.
\end{abstract}

%% file: sec/1_intro.tex
\section{Introduction}
Referring Video Object Segmentation (RVOS)~\cite{wu2022language,botach2022end,ding2023mevis} is an emerging task that aims to identify, track and segment the object in the video based on a natural language description.
Unlike traditional semi-supervised VOS models that require first-frame masks as the object prompt, RVOS models rely solely on text descriptions to segment the target. 
Considering its potential applications like video editing, growing efforts have been devoted to this field~\cite{luo2023soc,miao2023spectrum,ding2023mevis,he2024decoupling,liang2025referdino}. 
Recently, the advent of multi-modal large language models~\cite{liu2023visual,zhang2023video,jin2024chat} and segment anything models~\cite{kirillov2023segment,ravi2024sam2} has further accelerated this progress~\cite{bai2024one,yan2024visa,sa2va,zheng2024villa}.

\begin{figure}
  \centering
  \includegraphics[width=.44\textwidth]{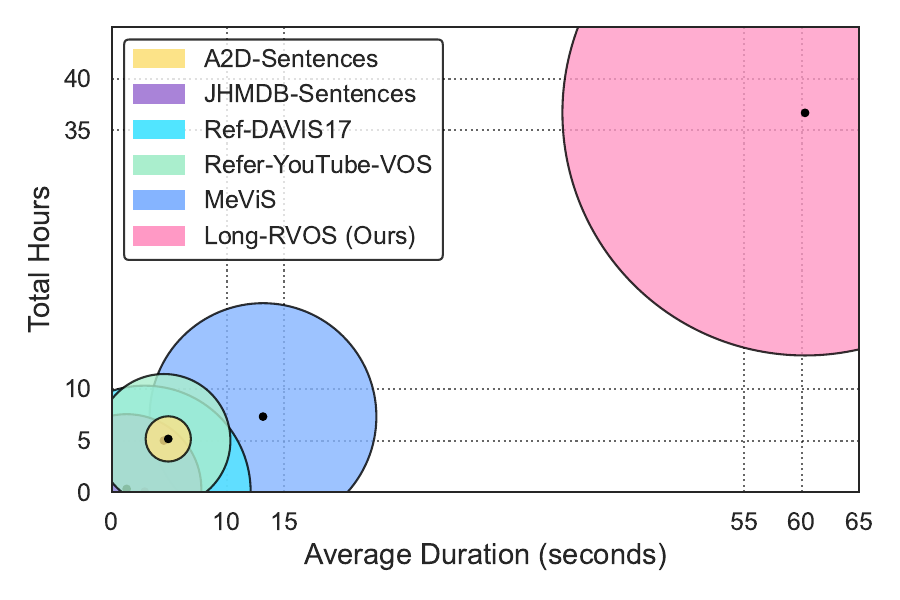}\vspace{-1em}
  \caption{Duration comparison of current RVOS datasets. The circle size indicates the number of frames.}\vspace{-1.5em}
  \label{fig:comp}
\end{figure}

\begin{figure*}[t]
    \centering
    \includegraphics[width=\textwidth]{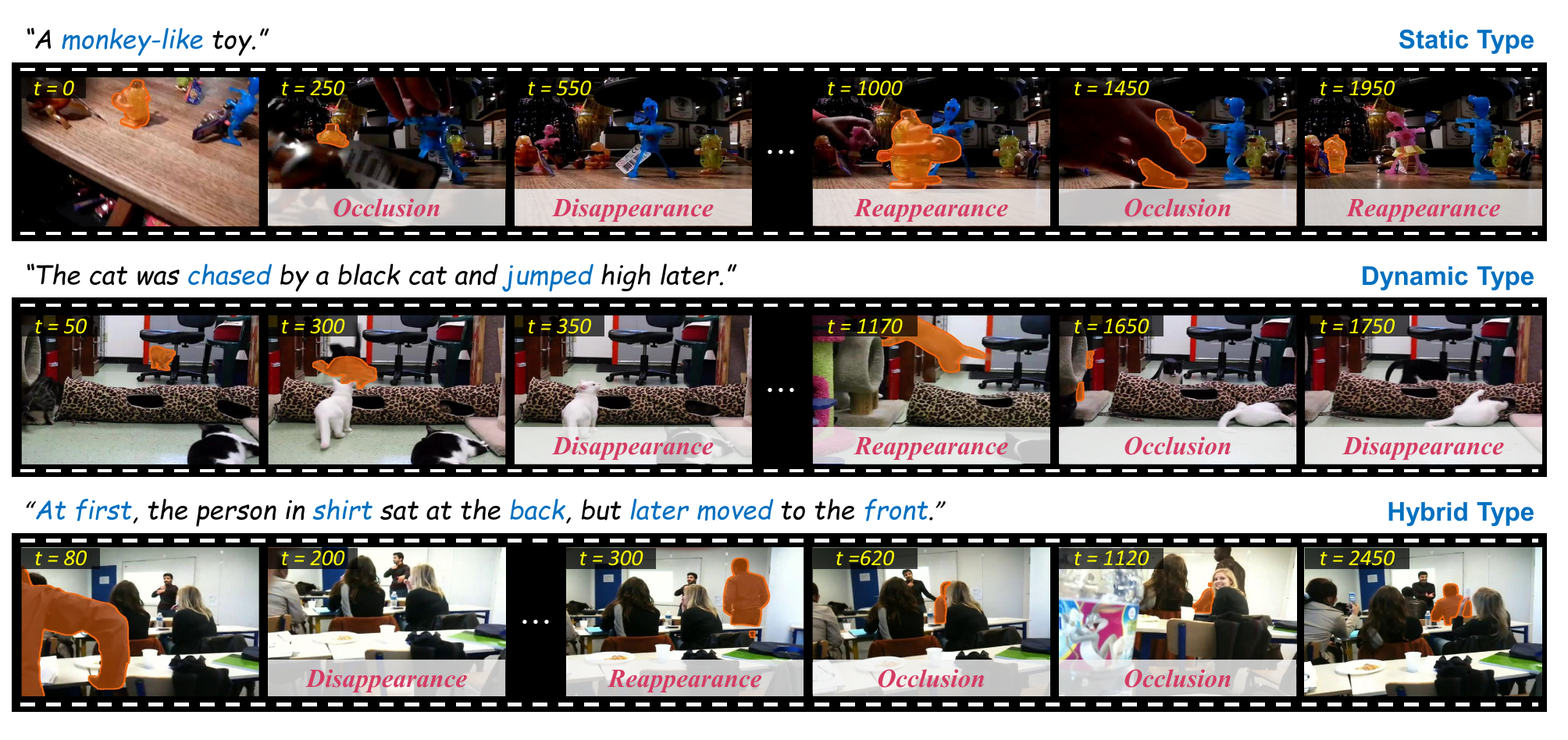}\vspace{-1em}
    \caption{Examples from Long-RVOS dataset, with frame indices displayed in the upper left, and selected objects masked in 
    \textcolor{mask_orange}{orange $\blacksquare$}.
    Long-RVOS contains extensive long-term videos, where the objects always undergo occlusion, disappearance-reappearance and shot changing. 
    In addition, the objects are annotated with three different types descriptions: \textit{Static}, \textit{Dynamic} and \textit{Hybrid}.}
    \label{fig:example}\vspace{-1em}
\end{figure*}

Despite these advances, current RVOS datasets~\cite{seo2020urvos,khoreva2019video,gavrilyuk2018actor,ding2023mevis} remain limited to short video clips lasting only a few seconds, with target objects clearly visible in most frames.
For state-of-the-art (SOTA) methods, in order to capture the target object accurately, it is inevitable to integrate as much spatiotemporal information as possible from the entire video.
However, when the video becomes longer, the number of distractors also increases accordingly, making it more challenging to perform sufficient spatiotemporal reasoning and capture the key information.
Especially in RVOS, many text descriptions (e.g., ``the cat jumping down'') only refer to a brief fragment in the video.
In other hand, due to the GPU memory limitation, existing methods~\cite{yan2024referred,liang2025referdino,lin2025glus,wang2025object} typically sample 4$\sim$8 frames per video for training, but use all the frames during inference.
As the video length increases, the gap between training and inference phases may become more pronounced.
Despite these concerns, due to the lack of a long-term RVOS benchmark, the exact challenges posed by longer videos remain unclear.

Another concern lies in the evaluation metrics. Existing RVOS benchmarks~\cite{seo2020urvos,khoreva2019video,gavrilyuk2018actor,ding2023mevis} typically evaluate performance by simply averaging the frame-wise segmentation metrics (e.g., $\cal{J}\&\cal{F}$). 
However, in real-world videos, the target objects do not appear in every frame, due to occlusion and constrained camera views.
Therefore, a robust RVOS model should exhibit a sound temporal consistency. 
This means it should not only segment the target when it is present, but also be able to predict its absence by outputting an empty mask.
However, this capability of temporal consistency can not be adequately reflected by existing metrics.

To address these gaps, this work proposes \textbf{Long-RVOS}, a large-scale benchmark for long-term video object segmentation. 
Long-RVOS is the first minute-level dataset in RVOS field, designed to tackle various realistic long-video challenges such as frequent occlusion, disappearance-reappearance and shot changing, as shown in \Cref{fig:comp} and \Cref{fig:example}.
Additionally, we introduce two new metrics for better evaluation of temporal consistency: 
$\mathrm{tIoU}$, which measures the temporal overlap between predicted and ground-truth mask sequences; and
$\mathrm{vIoU}$, which further measures the spatiotemporal volume overlap between them.
We benchmark 7 SOTA methods on Long-RVOS.
The results demonstrate that while notable progress has been achieved in existing short-term benchmarks, these SOTA models still significantly struggle in realistic long-term scenarios, in both frame-level segmentation and video-level temporal consistency.

To tackle the challenges posed by Long-RVOS, we present a baseline method \textbf{ReferMo}, which integrates additional motion frames to expand the temporal receptive field during training, and employs a local-to-global architecture to perceive both static attributes, short-term dynamics and long-term dependencies.
Specifically, ReferMo decomposes each video into a sequence of clips, each consisting of a high-resolution keyframe and multiple low-resolution motion frames. 
Then, it perceives the static appearance and short-term motion within local video clip, and captures the global target in long-term context via inter-clip interactions.
In this way, the temporal receptive field is expanded from multiple frames to multiple clips, but the training cost does not increase significantly.
Despite simplicity, ReferMo achieves significant improvements over existing RVOS approaches, serving a promising baseline for long-term referring video object segmentation.

To summarize, our contributions are two folds. 
\textbf{1)} We build Long-RVOS, the first large-scale long-term RVOS benchmark. In Long-RVOS, we provide explicit description types and introduce new metrics to enable more comprehensive evaluation.
\textbf{2)} We benchmark 7 SOTA approaches on Long-RVOS, and propose a promising baseline ReferMo to address the challenges in long-video scenarios.
These contributions establish a foundation for developing more robust RVOS models to handle realistic long videos.

% \textbf{(i)} We build Long-RVOS, the first large-scale long-term RVOS benchmark. In Long-RVOS, we provide explicit description types and introduce new metrics to enable more comprehensive evaluation.
% \textbf{(ii)} We benchmark 7 SOTA approaches on Long-RVOS, and propose a promising baseline ReferMo to address the challenges in long-video scenarios.
% These contributions establish a foundation for developing more robust RVOS models to handle the realistic long-term videos.

\begin{table*}[t]
  \caption{Statistical overview of representative RVOS datasets.
  Long-RVOS features the longest video duration and the most diverse object classes.
  Besides, Long-RVOS offers explicit text description types for finer-grained evaluation.
  }\label{tab:comparison}\vspace{-.5em}
  \setlength{\tabcolsep}{7pt}
  \resizebox{\textwidth}{!}{
  \begin{tabular}{lccccccccccc}
  \hline
  Dataset & Year & Videos & \makecell{Mean\\duration} & \makecell{Total\\duration} & \makecell{Mean\\frames} & \makecell{Masks} & Objects & \makecell{Object\\classes} & Text & \makecell{Text\\type} \\ \hline\hline
  A2D-Sentences~\cite{gavrilyuk2018actor} & 2018 & 3,782 & 4.9s & 5.2h & 3.2 & 58k & 4,825 & 6 & 6,656 & \ding{55}  \\ 
  JHMDB-Sentences~\cite{gavrilyuk2018actor} & 2018 & 928 & 1.3s & 0.3h & 34.3 & 32k & 928 & 1 & 928 & \ding{55} \\ 
  Ref-DAVIS17~\cite{khoreva2019video} & 2018 & 90 & 2.9s & 0.1h  & 69.0 & 14k & 205 & 78 & 1,544 & \ding{55} \\ 
  Refer-YouTube-VOS~\cite{seo2020urvos} & 2020 & \textbf{3,978} & 4.5s & 5.0h & 27.2 & 131k & 7,451 & 94 & 15,009 & \ding{55} \\ 
  MeViS~\cite{ding2023mevis}& 2023 & 2,006 & 13.2s & 7.3h & 79.0 & 443k & \textbf{8,171} & 36 & \textbf{28,570} & \ding{55} \\ \midrule
  \textbf{Long-RVOS} (ours) & 2025 & 2,193 & \textbf{60.3s} & \textbf{36.7h} & \textbf{361.7} & \textbf{2.1M} & 6,703 & \textbf{163} & 24,689 & \ding{51} \\ 
  \hline
  \end{tabular}}
  \vspace{-1em}
\end{table*}

%% file: sec/2_related_work.tex
\section{Related Works}
\textbf{RVOS Benchmarks.}
Given an object description, RVOS aims to identify, tracking and segment the referring object throughout the video.
This task was initially introduced by \citet{gavrilyuk2018actor} and \citet{khoreva2019video} in 2018, and has gradually become a popular topic in vision-language understanding.
\citet{gavrilyuk2018actor} built A2D-Sentences and JHMDB-Sentences, which focus on distinguishing different actors in a video through the descriptions about appearance and actions. 
\citet{khoreva2019video} built Ref-DAVIS17~\cite{khoreva2019video}, which covers more diverse object types.
Later, Ref-Youtube-VOS~\cite{seo2020urvos} was developed to further expand the benchmark scale in this field.
Recently, MeViS~\cite{ding2023mevis} was proposed to highlight the importance of motion understanding in RVOS task.
Despite the efforts, these benchmarks remain limited to short video clips lasting only a few seconds, with target objects clearly visible in most frames.
Besides, they also lack sufficient evaluation mechanisms to consider the models' specific capabilities in various aspects.
% These limitations hinder the advancement of the RVOS task towards more practical scenarios.

\vspace{.5em}\noindent\textbf{RVOS Approaches.}
Recent methods are primarily based on Transformer architecture, represented by MTTR~\cite{botach2022end} and ReferFormer~\cite{wu2022language}. 
For a consistent object identification across the frames, follow-up works~\cite{luo2023soc,tang2023temporal,han2023html,he2024decoupling} focus on integrating more object-level temporal information. 
ReferDINO~\cite{liang2025referdino} further improves the visual-language understanding by inheriting the object grounding capability of GroundingDINO~\cite{liu2024grounding}. 
Meanwhile, the recent emergence of segment anything models, i.e., SAM~\cite{kirillov2023segment} and SAM2~\cite{ravi2024sam2}, provides unique opportunity for downstream segmentation tasks.
Some frontier studies~\cite{yan2024visa,lin2025glus,cuttano2024samwise,bai2024one} explore to incorporate SAM and SAM2 into RVOS approaches.
% Sa2VA~\cite{sa2va,pixel_sail} combines SAM2 and LLaVA to achieve various language-guided segmentation.
For example, VideoLISA~\cite{bai2024one} incorporates large language models with SAM for reasoning video segmentation.
SAMWISE~\cite{cuttano2024samwise} integrate text prompts into SAM2 through trainable adapters.
% Due to GPU memory limitations, a common practice of current methods is sampling only several frames per video for training, which can lead to insufficient temporal understanding.
While these models achieve great progress in current short-video benchmarks, their abilities and robustness in handling real-world long videos is still unclear.

\vspace{.5em}\noindent\textbf{Long-term Video Understanding.} 
Real-world videos are always long, untrimmed, and involves multiple events.
To promote research into long-term video understanding, many large-scale benchmarks~\cite{wu2021towards,caba2015activitynet, mangalam2023egoschema,fu2024video} have been constructed.
However, these benchmarks are mainly constructed for video question answering and temporal action localization, containing only sparse annotations such as timestamps, action labels and captions.
To support object-level long-term understanding, some datasets including VidOR~\cite{shang2019annotating} and LaSOT~\cite{fan2019lasot} also provide dense annotations of bounding boxes. 
However, long-video datasets with pixel-level dense annotations are still very scarce.
Recently, LVOS~\cite{hong2023lvos} is built for long-term video object segmentation.
However, it is limited in scale and lacks text annotation.
In this work, we build Long-RVOS, the first large-scale benchmark for long-term video object segmentation, providing both pixel-wise annotations and diverse object descriptions.

%% file: sec/3_dataset.tex
\begin{figure*}[t]
  \centering
    \includegraphics[width=\linewidth]{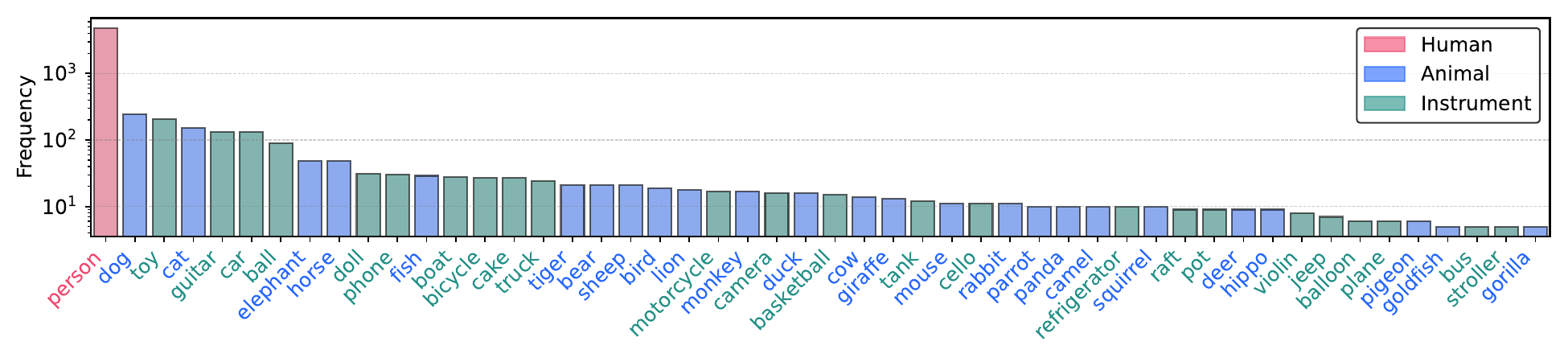}\vspace{-1.5em}
    \caption{Frequency distribution of the Top-50 object categories.}\label{fig:top50}
    \vspace{-1.5em}
\end{figure*}

\section{Long-RVOS}\label{sec:dataset}
\subsection{Video Collection}\label{sec:collection}
Previous RVOS datasets~\cite{gavrilyuk2018actor,khoreva2019video,seo2020urvos,ding2023mevis} were typically constructed by providing text annotations on their corresponding VOS datasets (e.g., DAVIS17~\cite{pont20172017}, YouTube-VOS-2019~\cite{xu2018youtube} and MOSE~\cite{ding2023mose}). However, the existing long-term VOS datasets like LVOS~\cite{hong2023lvos} are limited in scale (only 720 videos), and most videos feature only one object target.
Therefore, in order to establish a large-scale and diverse RVOS benchmark, we bypass the existing VOS datasets and turn to multi-source long video datasets.
Specifically, we integrate three long-video datasets: TAO~\cite{dave2020tao}, VidOR~\cite{shang2019annotating}, and Ego-Exo4D~\cite{grauman2024ego}.
Additionally, TAO is a federated dataset combining multiple sources like Charades~\cite{sigurdsson2016hollywood}, LaSOT~\cite{fan2019lasot}, ArgoVerse~\cite{chang2019argoverse}, AVA~\cite{gu2018ava}, YFCC100M~\cite{thomee2016yfcc100m}, BDD-100K~\cite{yu2020bdd100k}, and HACS~\cite{zhao2019hacs}.
These datasets typically provide bounding box annotations on sparse frames.
Then, we select videos and objects based on the following criteria:
\begin{itemize}[leftmargin=*]
  \item The video duration exceeds 20 seconds.
  \item Objects that belong to background, ambiguous or unknown categories are excluded.
  \item Each selected video must contain more than two valid objects, and at least one object is not continuously visible. 
\end{itemize}
With these criteria, we have initially collected over 3K videos and 8K objects as candidates.
After careful inspections on quality, we finally select 2,193 videos and 6,703 objects to build Long-RVOS.

\subsection{Dataset Annotation.}
\textbf{Text Annotation.}
We develop an online platform for annotating object descriptions. 
This platform randomly samples a video from our dataset and displays it, with all target objects highlighted by bounding boxes.
To ensure the diversity of annotations, each video can be sampled repeatedly at most three times.
The annotators consisting of 20 college students are asked to watch the videos and provide the following three types of descriptions for each object:
\begin{itemize}[leftmargin=*]
  \item \textbf{Static type} includes appearance (e.g., colors and shapes), relative position (e.g., ``the left cat''), and environmental context (e.g., ``on the grass'').
  \item \textbf{Dynamic type} includes motions, changes over time (e.g., in position or state) and interactions with other entities (e.g., ``the cat chasing a mouse'').
  \item \textbf{Hybrid type} integrates both static and dynamic attributes to provide comprehensive object cues.
\end{itemize}
The key annotation principle is that every single description, regardless of type, must clearly distinguish the target object from others.
For objects that cannot be distinguished by only static or dynamic attributes, the corresponding type of annotation can be skipped.
After this annotation phase, we have collected over 30K text descriptions.
These annotations and the corresponding videos are then sent to a validation team formed by three experts for quality verification.
Any descriptions that violate our principle are directly removed.
Besides, we do not use techniques like synonym replacement to artificially scale up the text annotations, keeping the dataset clear and authentic to support reliable RVOS training.
Finally, we gather 24,689 high-quality descriptions for building Long-RVOS.

\begin{figure}
  \centering
  \includegraphics[width=0.45\textwidth]{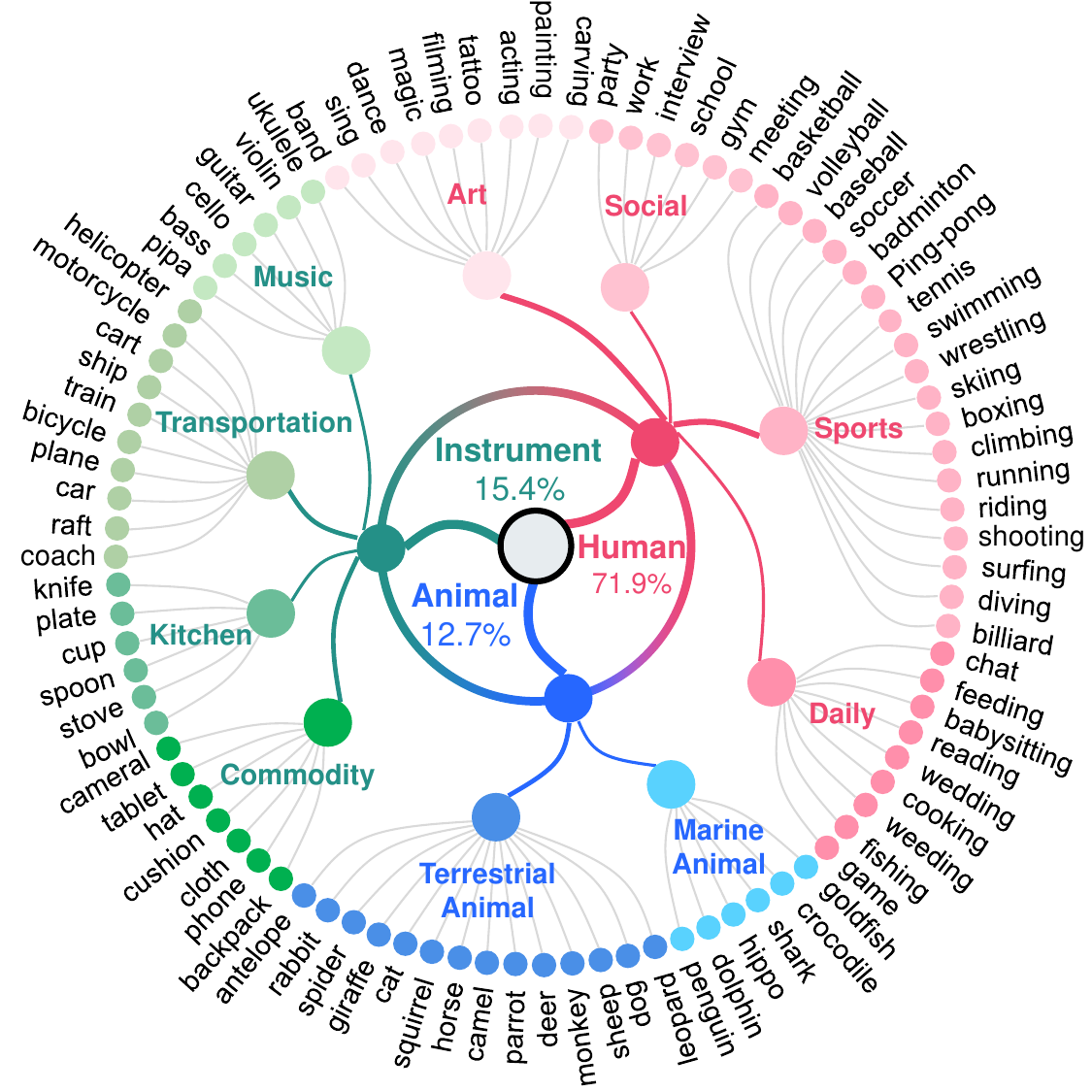}\vspace{-.5em}
  \caption{Overview of objects and scenes in Long-RVOS.}
  \label{fig:type}\vspace{-1em}
\end{figure}

\noindent\textbf{Mask Annotation.}
Our source datasets~\cite{dave2020tao,shang2019annotating,grauman2024ego} have provided sparse bounding-box annotations.
For each object, we segment the video into clips based on the annotated frames. 
Then, we utilize SAM2~\cite{ravi2024sam2}, the state-of-the-art VOS model, to track the objects within each clip and produce high-quality masks, by regarding the annotated bounding box as the first-frame prompt. 
% However, due to the complex video scenes like occlusion and motion, these initial mask sequences can be inaccurate.
To ensure annotation quality, we conduct an iterative \textit{check–correct} workflow.
Specifically, the validation team checks every object's mask separately in the video, and marks the objects with inaccurate annotations.
To facilitate the correction process, we develop an interactive annotation tool based on SAM2.
This tool loads a marked object each time and visualizes its masks in the video.
Nine annotators use our tool to refine the masks with point or box prompts, and remove masks from object-absent frames.
The corrected results are then returned to the checking queue, and this \textit{check–correct} loop repeats until all mask annotations are qualified.
% Totally, 261 objects are corrected during this process.

% \begin{figure}[h!]
%   \centering
%   \begin{minipage}[b]{0.42\textwidth}
%       \adjustbox{width=\textwidth}{\includegraphics{fig/word.pdf}}
%   \end{minipage}
%   \hfill
%   \begin{minipage}[b]{0.56\textwidth}
%       \adjustbox{width=\textwidth}{\includegraphics{fig/type.pdf}}
%   \end{minipage}
%   \caption{}
%   \label{fig:intro}
% \end{figure}

\subsection{Dataset Statistics}\label{sec:dataset_stat}
A detailed comparison with five existing RVOS datasets is shown in \Cref{tab:comparison}.
Notably, Long-RVOS offers significantly longer video duration than existing datasets.
In addition, it contains the largest number of object classes and mask annotations.
The large scale of Long-RVOS supports comprehensive training and evaluation of RVOS models.
% More attributes of Long-RVOS are described below.

\begin{figure}[t]
  \begin{subfigure}[t]{0.49\columnwidth}
    \includegraphics[width=\linewidth]{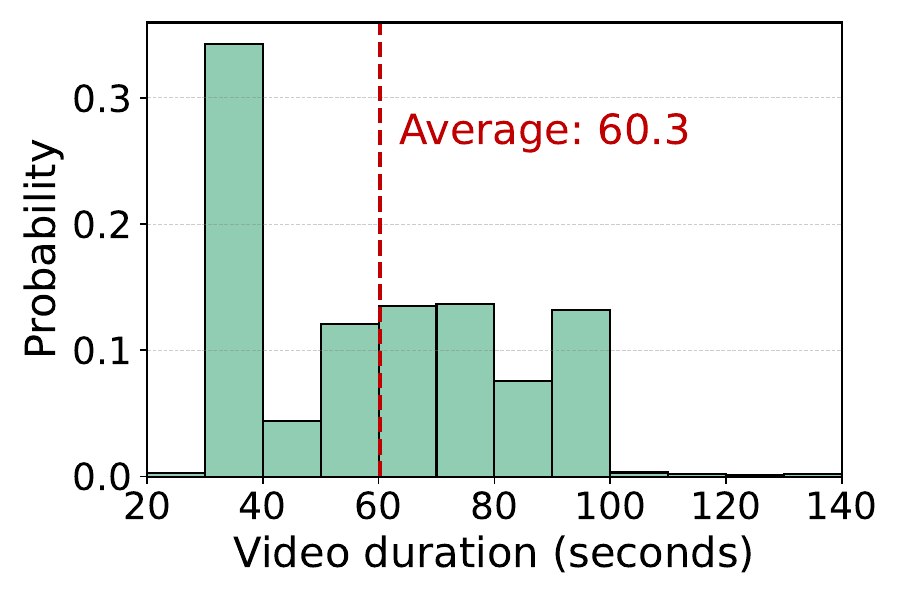}\vspace{-.5em}
    \caption{Distribution of video duration.}\vspace{.5em}
  \end{subfigure}\hfill
  \begin{subfigure}[t]{0.49\columnwidth}
    \includegraphics[width=\linewidth]{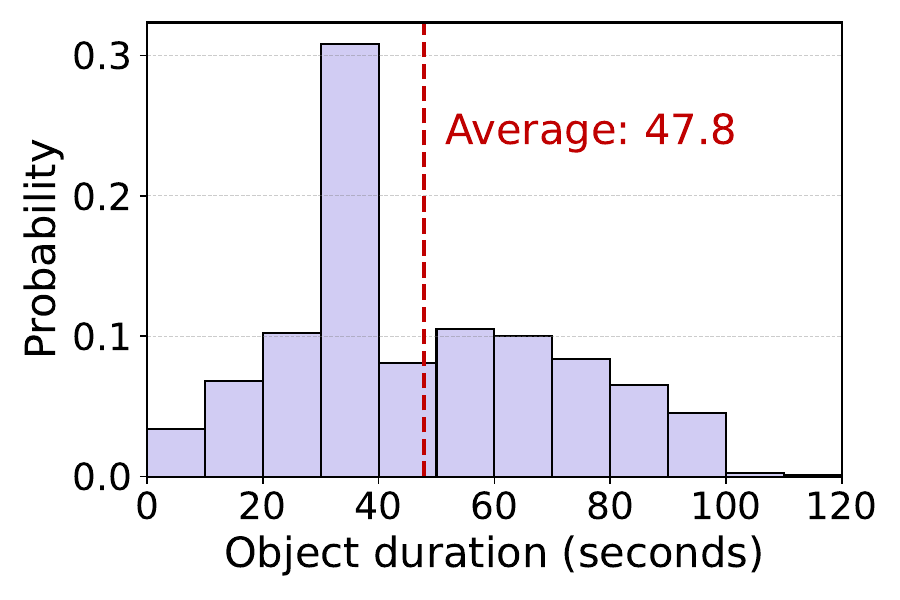}\vspace{-.5em}
    \caption{Distribution of object duration.}\vspace{.5em}
  \end{subfigure}
  \begin{subfigure}[t]{0.49\columnwidth}
    \includegraphics[width=\linewidth]{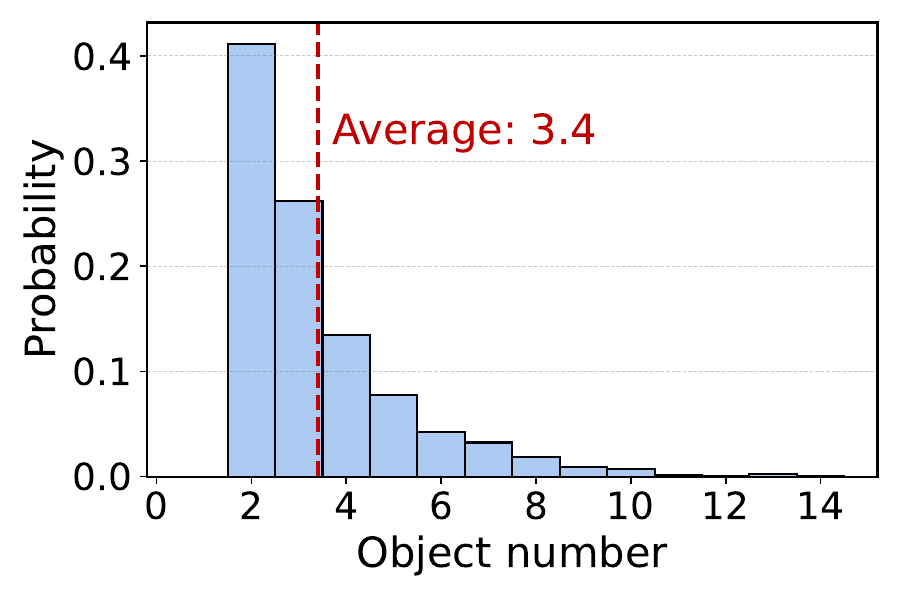}\vspace{-.5em}
    \caption{Number of objects per video.}
  \end{subfigure}\hfill
  \begin{subfigure}[t]{0.49\columnwidth}
    \includegraphics[width=\linewidth]{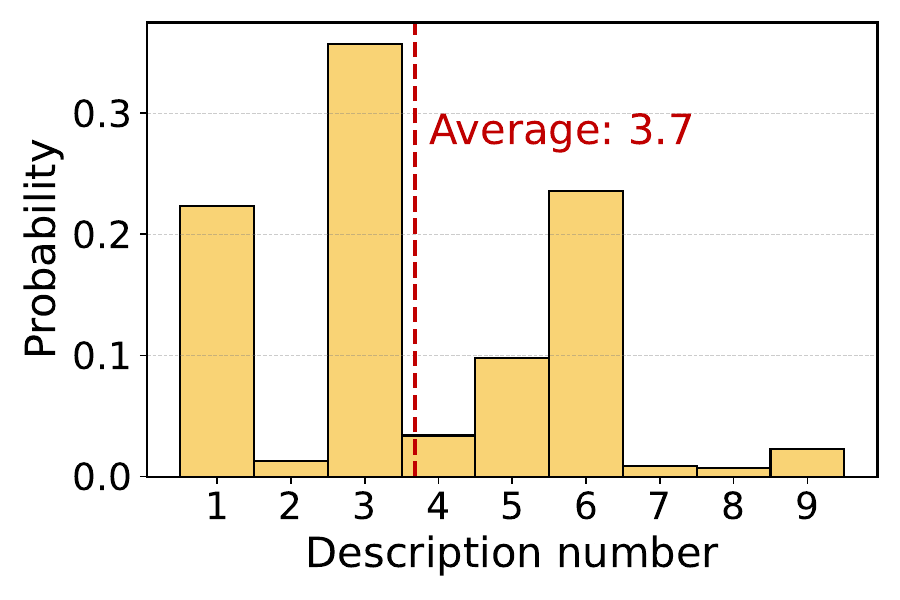}\vspace{-.5em}
    \caption{Description number per object.}~\label{fig:dec_num}
  \end{subfigure}\vspace{-1.5em}
  \caption{Representative statistics of Long-RVOS.}
  \label{fig:stat}\vspace{-1.5em}
\end{figure}

\noindent\textbf{Diverse Objects and Scenes.} 
Long-RVOS is constructed by integrating multiple sources of video datasets, achieving a wide variety of objects and scenes, as illustrated in Figures \ref{fig:top50} and \ref{fig:type}.
These sources include indoor videos from Charades~\cite{sigurdsson2016hollywood}, outdoor videos from LaSOT~\cite{fan2019lasot}, movie scenes from AVA~\cite{gu2018ava}, egocentric videos from Ego-Exo4D~\cite{grauman2024ego}, and more diverse videos from other datasets~\cite{shang2019annotating,thomee2016yfcc100m,zhao2019hacs}.
In total, Long-RVOS contains 163 object categories, significantly surpassing the existing RVOS datasets.
While Long-RVOS primarily focuses on human instances (71.9\%), it also covers a diverse range of animals (12.7\%) and instruments (15.4\%).
In \Cref{fig:stat}, we present further statistics on the videos and objects in Long-RVOS.
Notably, the object number of each video spans from 2 to 14, preventing over-reliance on the most salient object and highlighting text-guided segmentation.
Such extensive visual diversity ensures that models are tested against a wide array of complex, real-world scenarios.

\begin{table}[t]
  \centering
  \footnotesize
  \caption{Distribution of description types.}
  \label{tab:type_stat}\vspace{-1em}
  \setlength{\tabcolsep}{12pt}
    \resizebox{.95\linewidth}{!}{
    \begin{tabular}{r|ccc}
      \hline
        Type & \it{Static} & \it{Dynamic} & \it{Hybrid} \\
      \hline
        Percentage & 35.03\% & 32.45\% & 32.52\% \\
      \hline
    \end{tabular}}\vspace{-.5em}
\end{table}

% \begin{figure}[t]
%     \centering
%     \includegraphics[width=\linewidth]{fig/bar1.pdf}\vspace{-1em}
%     \caption{Distribution of description types.}
%     \label{tab:type_stat}\vspace{-1.5em}
% \end{figure}

\begin{figure}[t]
    \centering
    \includegraphics[width=.9\linewidth]{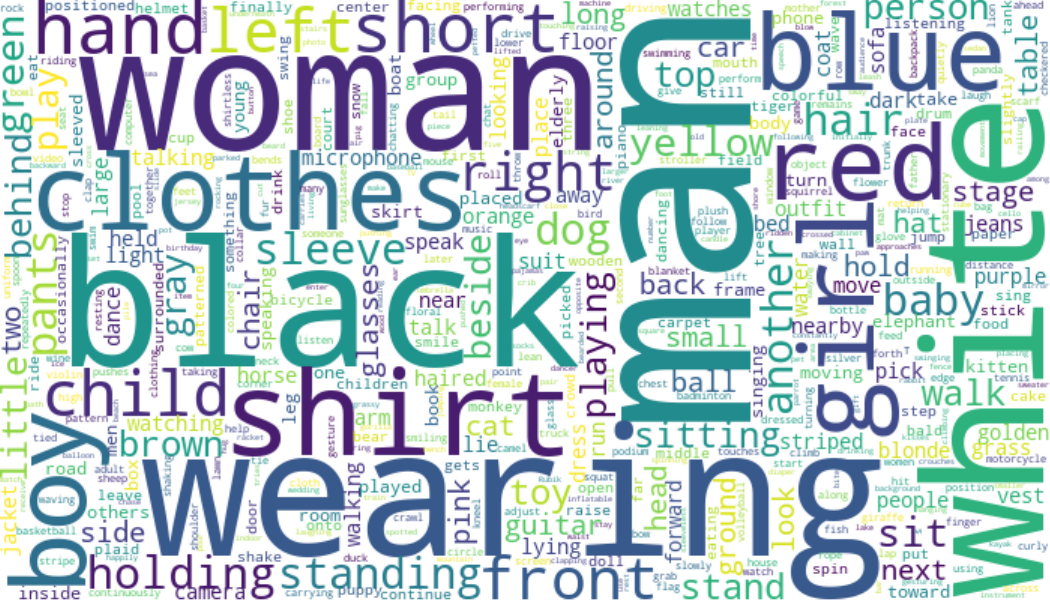}\vspace{-.5em}
    \caption{Wordcloud of descriptions.}\label{fig:wordcloud}\vspace{-1.5em}
\end{figure}

\noindent\textbf{Diverse Descriptions.}
In real-world applications, user queries are always unpredictable. They might refer to salient attributes or instantaneous actions.
To enable more comprehensive evaluation of model capabilities, Long-RVOS introduces three distinct types of text descriptions --- \textit{Static}, \textit{Dynamic}, and \textit{Hybrid}.
By explicitly categorizing these types, Long-RVOS prevents evaluation bias toward specific attribute cues (e.g., colors or positions), ensuring a fair and robust assessment.
As shown in \Cref{tab:type_stat}, Long-RVOS maintains a balanced distribution among these three description types.
In addition, \Cref{fig:dec_num} illustrates that the description number for each object can vary from 1 to 9.
These properties encourage comprehensive learning of diverse object attributes.
We also present the wordcloud of Long-RVOS is in \Cref{fig:wordcloud}.
Together, the diversity in both visual content and textual descriptions establishes Long-RVOS as a truly comprehensive benchmark for the training and evaluation of long-form RVOS models.

% \textbf{Dataset Attribute}

\subsection{Evaluation Metrics}
Previous RVOS benchmarks tend to evaluate model performance with the frame-wise spatial metrics, such as $\cal{J}\&\cal{F}$.
Here, $\cal{J}$ denotes the Intersection-over-Union (IoU) between the predicted and ground-truth masks, $\cal{F}$ measures the contour accuracy, and $\cal{J}\&\cal{F}$ is their average over all the frames.
However, these metrics focus solely on the per-frame segmentation quality, neglecting the temporal consistency.
A robust RVOS model should accurately segment the target when it is present and correctly output an empty mask when it is absent.
Inspired by the field of spatiotemporal video grounding~\cite{zhang2020does,tang2021human}, we additionally introduce two new metrics, $\mathrm{tIoU}$ and $\mathrm{vIoU}$, in Long-RVOS to individually evaluate the temporal and spatiotemporal performance.

Formally, let $\hat{M}_t,M_t \in \{0,1\}^{H\times W}$ denote the predicted and ground-truth masks at $t$-th frame, respectively, where $t \in [1, T]$.
The frame-index sets of non-empty masks are defined as $\hat{\mathcal{T}} = \{ t \mid \| \hat{M}_t \|_0 > 0 \}$ (for predictions) and $\mathcal{T} = \{ t \mid \| M_t \|_0 > 0 \}$ (for the ground-truth), where the $\ell_0$-norm $\|\cdot\|_0$ denotes the count of non-zero elements.
Then, $\mathrm{tIoU}$ is obtained by computing their IoU as follows:
\begin{equation}
  \mathrm{tIoU} = \frac{T_i}{T_u}, \quad \text{where } T_i = \hat{\mathcal{T}} \cap \mathcal{T} \text{ and } T_u = \hat{\mathcal{T}} \cup \mathcal{T},
\end{equation}
and $\mathrm{vIoU}$ computes the volume IoU between predicted and ground-truth mask sequences: 
\begin{equation}
  \mathrm{vIoU} = \frac{1}{T_u} \sum_{t\in T_i} \mathcal{J}_t, \quad \text{where } \mathcal{J}_t = \frac{\hat{\mathcal{M}}_t \cap \mathcal{M}_t}{\hat{\mathcal{M}}_t \cup \mathcal{M}_t}.
\end{equation}

By combining the spatial metric $\cal{J}\&\cal{F}$, temporal metric $\mathrm{tIoU}$ and spatiotemporal metric $\mathrm{vIoU}$, Long-RVOS establishes a rigorous evaluation protocol for RVOS research.

%% file: sec/4_method.tex
\begin{figure*}[t]
  \centering
  \includegraphics[width=\textwidth]{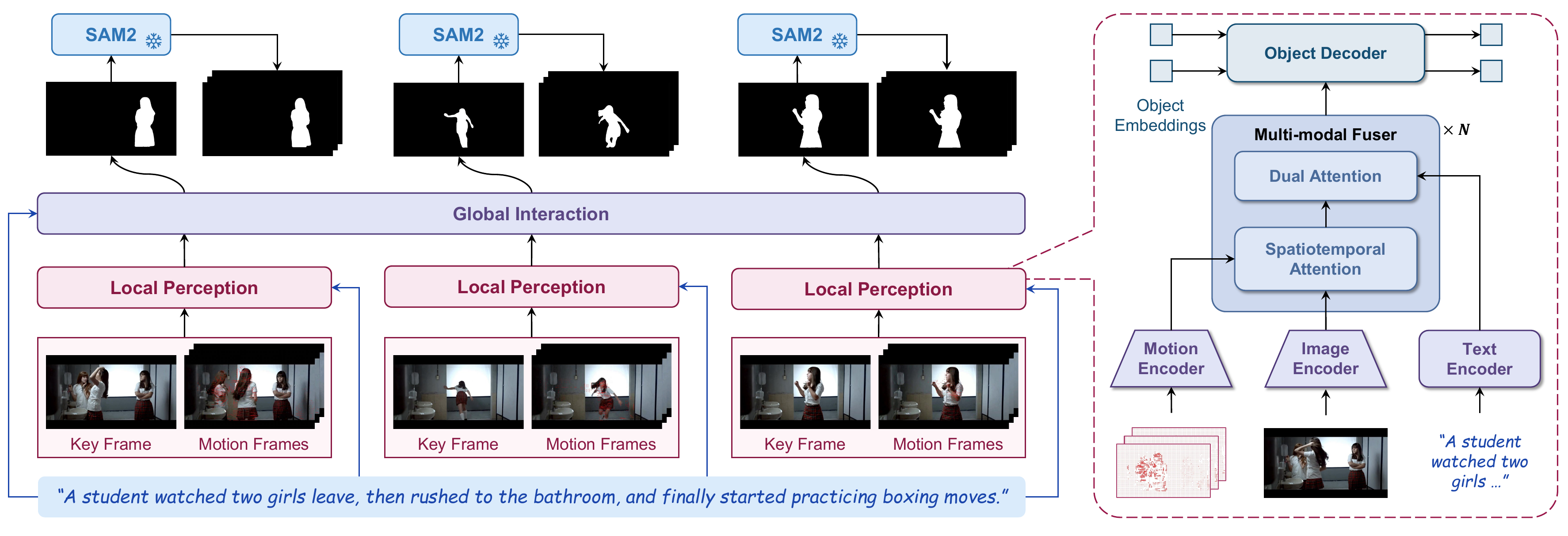}\vspace{-1em}
  \caption{Overview of ReferMo. A video is decomposed into clips (keyframe + motion frames). ReferMo perceives the static attributes and short-term motions within each clip, then aggregates inter-clip information capture the global target.
  Notably, ReferMo is supervised by only keyframe masks, and SAM2 is only used at inference for target tracking in subsequent frames.
  }\vspace{-1em}
  \label{fig:model}
\end{figure*}

\section{ReferMo: A Baseline Approach}\label{sec:baseline}
As illustrated in \Cref{fig:model}, ReferMo decomposes the video into a sequence of clips, each consisting of a high-resolution keyframe and subsequent low-resolution motion frames.
Then, it perceives the static appearance and short-term motion within local video clip, and captures the object target in long-term context by integrating the cross-clip information. 
Critically, ReferMo only predicts target masks over the keyframes, and the masks on the remain frames are generated by a pretrained mask tracker (e.g., SAM2~\cite{ravi2024sam2}).
In this way, ReferMo achieves a trade-off between training costs and long-term understanding without processing a large number of high-resolution frames.

\subsection{Video Decomposition}
Typically, a long-term video is composed of multiple shots, and the video frames within each shot often show significant temporal redundancy.
This redundancy can be efficiently described by motion information to reduce the frame-by-frame computations.
Inspired by Video-LaVIT~\cite{jin2024video}, we employ the MPEG-4~\cite{le1991mpeg} compression technique to extract keyframe and motion information from the videos.
More sophisticated (but expensive) keyframe selection strategies~\cite{wu2019adaframe,wang2024videoagent} can also be explored, but they are not the primary focus of this work.
In MPEG-4, a video is decomposed into multiple clips, where each clip consists of a keyframe $\mathcal{I} \in \mathbb{R}^{H\times W\times 3}$ and the motion vectors $\mathcal{M} \in \mathbb{R}^{T\times\frac{H}{16}\times\frac{W}{16}\times2}$ of its subsequent $T$ frames.
Unlike the dense optical flow, these motion vectors can be directly extracted during the compressed video decoding process, making them well-suited for processing large-scale, long-term videos.
More details of motion extraction process are provided in the supplementary.

% Formally, each frame in a video clip is partitioned into $16\times16$ non-overlapping macroblocks. The motion vector $\vec{m}$ of the $t$-th frame is estimated as follows:
% \begin{equation}
%   \vec{m}(p,q) = \arg\min_{i,j} \Vert I_t(p,q) - I_{t-1}(p-i,q-j) \Vert,
% \end{equation}
% where $I(p,q)$ indicates the pixel values of the macroblock at location $(p,q)$, and $(i,j)$ is the coordinate offset between the centers of the two macroblocks.

% Unlike traditional methods that use the expensive dense optical flow for motion estimation, we resort to motion vectors since they can be directly extracted during the compressed video decoding process.

\subsection{From Local Perception to Global Interaction}

% \begin{figure}
%   \centering
%   \includegraphics[width=0.3\textwidth]{fig/fusion.pdf}
%   \caption{Overview of local perceiver.}
%   \label{fig:local}
% \end{figure}

Different from the previous RVOS methods~\cite{luo2023soc,yan2024referred,liang2025referdino} that perform vision-language fusion on each single frame, we introduce motion representations to enable clip-level vision-language fusion.
For each video clip, as shown in the right part of \Cref{fig:model}, the local perceiver encodes the text, keyframe and motion information through three separate encoders, and then employs a multi-modal fuser to progressively aggregate these information for clip-level object extraction. 
By collecting the objects across video clips, we perform global temporal interaction to enable consistent object prediction and long-term temporal understanding.

\vspace{.5em}\noindent\textbf{Motion Encoder.}
The motion vectors are first embeded into a $d$-dimensional space via a linear projector.
Then, the motion encoder performs self-attention separately along the spatial and temporal dimensions to extract the spatiotemporal motion features $M\in \mathbb{R}^{T\times \frac{H}{16}\times \frac{W}{16} \times d}$. Notably, we implement the spatial attention as deformable attention due to the large number of spatial tokens.

% Given the motion vectors of $\mathbb{R}^{T\times \frac{H}{8}\times \frac{W}{8}\times 2}$ of a video segment, we adopt a linear projector to transform them into a $d$-dimensional embedding .
% These motion embeddings, added with the spatial and temporal position embeddings, are then fed into a spatial attention layer, a temporal attention layer and a feedforward layer to encode the spatiotemporal features.
% Specially, we implement the spatial attention as deformalble attention due to the large number of spatial tokens.
% In addition, modern image encoder tends to generates the multi-scale image features with spatial shapes of $\frac{1}{8}$, $\frac{1}{16}$, $\frac{1}{32}$ and $\frac{1}{64}$. 
% To match the spatial shapes of image features, we also employ a sequence of convolution kernels over the motion features to generate multi-scale features.

\vspace{.5em}\noindent\textbf{Image-Motion Fusion.}
Modern image encoders (e.g., Swin Transformer~\cite{liu2021swin}) typically output multi-scale feature maps $I_i \in \mathbb{R}^{H_i\times W_i\times d}$, $i\in[1,4]$.
To match these spatial resolutions, we adopt a series of spatial convolutions with specific strides over the motion features $M$ to produce multi-scale motion features $M_i \in \mathbb{R}^{T \times H_i \times W_i \times d}$.
At each scale $i$, we treat the keyframe feature $I_i$ as \textit{query} and perform cross-attention along the temporal dimension to aggregate $M_i$ into $\widetilde{M}_i \in \mathbb{R}^{H_i\times W_i\times d}$.
To avoid undesired motion noise, we fuse the keyframe and motion features via the spatial-aware and channel-aware gating mechanisms:
{
  \addtolength{\abovedisplayskip}{-3pt}
  \addtolength{\belowdisplayskip}{-3pt}
\begin{equation}
  M_i^* = (\underbrace{\sigma(I_i \cdot W_{down}^I)}_{\text{Spatial Gate}} \odot (\widetilde{M}_i \cdot W_{down}^M))\cdot W_{up}, \\
\end{equation}}
\begin{equation}
  F_i = I_i + \smallunderbrace{\gamma_i}{\mathclap{\text{Channel Gate}}} \odot \max(M_i^{*},0)^{2},
\end{equation}
where $W_{down}^I, W_{down}^M \in \mathbb{R}^{d\times r}$ indicate the low-rank projectors that compress the features to a lower dimension $r$, and $W_{up} \in \mathbb{R}^{r\times d}$ resorts the dimension. $\sigma$ denotes Sigmoid function and $\odot$ denotes Hadamard product. $\gamma \in \mathbb{R}^d$ is a learnable vector to modulate the channel-wise weights. 

\vspace{.5em}\noindent\textbf{Vision-Language Fusion.} We use dual cross-attention~\cite{li2022grounded,liu2024grounding} for deep vision-language fusion. Formally, given the clip-level vision features $F \in \mathbb{R}^{N\times d}$ and the language features $E \in \mathbb{R}^{L\times d}$, where $N$ and $L$ individually denote their token number, we derive the cross-modal enhanced vision features $\widetilde{F}$ and language features $\widetilde{E}$ as follows:
\begin{equation}
\left\{
\begin{aligned}
&\widetilde{F} = \mathrm{Softmax}(\mathcal{A}) \cdot E, \\
&\widetilde{E} = \mathrm{Softmax}(\mathcal{A}^\top) \cdot F,
\end{aligned}
\right.
\qquad \text{where } \mathcal{A} = \frac{F E^\top}{\sqrt{d}}.
\end{equation}
% {
%   \addtolength{\abovedisplayskip}{-3pt}
%   \addtolength{\belowdisplayskip}{-3pt}
% \begin{equation}
%   \mathcal{A} = \frac{F \cdot E^\top}{\sqrt{d}}, \quad \widetilde{F} = \mathrm{Softmax}(\mathcal{A}) \cdot E, \quad \widetilde{E}= \mathrm{Softmax}(\mathcal{A}^\top) \cdot F.
% \end{equation}}
For simplicity, the linear projections for multi-head attentions are omitted. The output features $\widetilde{F}$ and $\widetilde{E}$ are then fed into the object decoder to extract object features.

\vspace{.5em}\noindent\textbf{Global Interaction.}
To enable consistent object prediction and long-term temporal understanding, we collect the object features across video clips to perform global temporal interactions. Following ReferDINO~\cite{liang2025referdino}, we use the Hungarian algorithm~\cite{kuhn1955hungarian} to align the objects clip-by-clip. Then, we perform temporal self-attention over the aligned object features to achieve global modeling. 
For better modality alignment, we also infuse the language information $\widetilde{E}$ into the object features through a cross-attention layer.
Finally, the interacted object features are sent to the segmentation head for generating instance masks.
Note that these masks are only predicted for the key frame within each clip, serving as object anchors for SAM2's mask propagation in subsequent frames. More details are present in the supplementary.

%% file: sec/5_experiment.tex
\begin{table*}[t]
  \setlength{\tabcolsep}{5pt}
  \footnotesize
  \centering
  \caption{Comparison on Long-RVOS test set. FPS is estimated at 360P on Nvidia A6000 GPUs, excluding the video loading time.}\vspace{-1em}
  \resizebox{\textwidth}{!}{
    \begin{tabular}{l |ccc |ccc |ccc |ccc |c}
      \hline
      \multirow{2}{*}{Method} & \multicolumn{3}{c|}{Static} & \multicolumn{3}{c|}{Dynamic} & \multicolumn{3}{c|}{Hybrid} & \multicolumn{3}{c|}{Overall} & \multirow{2}{*}{FPS} \\ 
      & $\mathcal{J}\&\mathcal{F}$ & $\mathrm{tIoU}$ & $\mathrm{vIoU}$ & $\mathcal{J}\&\mathcal{F}$ & $\mathrm{tIoU}$ & $\mathrm{vIoU}$ & $\mathcal{J}\&\mathcal{F}$ & $\mathrm{tIoU}$ & $\mathrm{vIoU}$ & $\mathcal{J}\&\mathcal{F}$ & $\mathrm{tIoU}$ & $\mathrm{vIoU}$ & \\
      \hline\hline
      {\scriptsize\color{cvprblue}\textit{Without SAM / SAM2}} &&&&&&&&&&&& \\
      SOC~\cite{luo2023soc} {\scriptsize\color{gray}NeurIPS'23} & 39.3 & 71.8 & 33.9 & 38.8 & 73.2 & 34.2 & 37.7 & 71.9 & 32.5 & 38.6 & 72.3 & 33.5 & 53.8 \\
      MUTR~\cite{yan2024referred} {\scriptsize\color{gray}AAAI'24} & 42.8 & 72.6 & 38.7 & 41.2 & 73.5 & 37.7 & 42.4 & 72.3 & 38.1 & 42.2 & 72.8 & 38.2 & 20.4 \\
      ReferDINO~\cite{liang2025referdino} {\scriptsize\color{gray}ICCV'25} & 50.9 & 73.6 & 46.0 & 45.4 & 73.8 & 41.5 & 48.7 & \textbf{73.1} & 44.0 & 48.4 & 73.5 & 43.9 & 46.4 \\
      \hline
      {\scriptsize\color{cvprblue}\textit{With SAM / SAM2}} &&&&&&&&&&&& \\
      VideoLISA~\cite{bai2024one} {\scriptsize\color{gray}NeurIPS'24} & 17.7 & 65.8 & 14.0 & 12.7 & 71.4 & 5.3 & 11.5 & 70.0 & 4.6 & 14.0 & 69.0 & 8.1 & 6.6 \\
      GLUS~\cite{lin2025glus} {\scriptsize\color{gray}CVPR'25} & 25.2 & 61.8 & 21.6 & 27.2 & 62.7 & 23.9 & 24.8 & 60.3 & 20.6 & 25.7 & 61.6 & 22.0 & 3.6 \\
      SAMWISE~\cite{cuttano2024samwise} {\scriptsize\color{gray}CVPR'25} & 41.3 & 65.5 & 31.3 & 40.4 & 67.6 & 31.3 & 41.0 & 66.9 & 30.8 & 40.9 & 66.6 & 31.1 & 7.0 \\
      RGA3~\cite{wang2025object} {\scriptsize\color{gray}ICCV'25} & 22.1 & 59.8 & 16.9 & 23.4 & 61.0 & 19.0 & 22.2 & 59.2 & 16.7 & 22.5 & 60.0 & 17.5 & 8.7 \\
      \hline
      \rowcolor{cvprblue!10} \textbf{ReferMo} (Ours) & \textbf{55.8} & \textbf{73.6} & \textbf{47.5} & \textbf{49.3} & \textbf{74.2} & \textbf{42.4} & \textbf{53.3} & 72.9 & \textbf{45.4} & \textbf{52.9} & \textbf{73.6} & \textbf{45.2} & 52.5 \\
      \hline
    \end{tabular}}\vspace{-1em}
  \label{tab:test} 
\end{table*}

\section{Experiments}\label{sec:exp}
\subsection{Experiment Setup}\label{sec:setting}
\textbf{Dataset Split.}
Long-RVOS is a large-scale dataset containing 2,193 videos and 24,689 descriptions. It is split into three subsets:
a training set of 1,855 videos and 20,722 descriptions, a validation set of 112 videos and 1,326 descriptions, and a test set of 226 videos and 2,641 descriptions.

\noindent\textbf{Evaluation Metrics.}
We use three kinds of evaluation metrics: the spatial metric $\cal{J}\&\cal{F}$, the temporal metric $\mathrm{tIoU}$ and the spatiotemporal metric $\mathrm{vIoU}$.
Long-RVOS provides three types of descriptions: \textit{Static}, \textit{Temporal} and \textit{Hybrid}.
We report performance for each type separately and overall.
We also recommend reporting FPS because efficiency is a major concern for long-video processing.

\noindent\textbf{Implementation Details.}\label{sec:detail}
We follow the default hyper-parameter settings of ReferDINO~\cite{liang2025referdino} and use Swin-Tiny as the backbone.
For SAM2~\cite{ravi2024sam2}, we use the $\mathrm{sam2.1\_hiera\_large}$ version.
In MPEG-4~\cite{le1991mpeg}, each video clip typically consists of a keyframe and the motion vectors for up to 11 subsequent frames.
During training, we randomly sample 6 clips and use 3-frame motion vectors.
The input frames are resized to have the longest side of 640 pixels and the shortest side of 360 pixels during both training and evaluation.
Following MeViS~\cite{ding2023mevis}, we do not use image segmentation datasets (e.g., RefCOCO/+/g~\cite{kazemzadeh2014referitgame, mao2016generation}) for pretraining.
We train ReferMo on Long-RVOS dataset for 6 epochs, which take 24 hours on 8 Nvidia A6000 GPUs.

\subsection{Benchmark Results}\label{sec:main_result}
\textbf{Overall Comparison.}
We compare ReferMo with 7 recent RVOS methods on Long-RVOS.
For a fair comparison, all the models are trained solely on Long-RVOS, with no external segmentation datasets used.
As demonstrated in \Cref{tab:test}, realistic long-video scenarios remain a significant challenge for current RVOS models.
While the SAM2-based methods~\cite{lin2025glus,cuttano2024samwise,wang2025object} achieve SOTA performance on existing short-term benchmarks, they significantly struggle in Long-RVOS. 
This suggests that their improvements may primarily stem from SAM2's superior tracking and segmentation capabilities, rather than better language-object understanding.
As videos grow longer and more complex, it becomes more challenging to maintain accurate video-language reasoning and consistently distinguish the objects, which leads to their performance degradation.
This issue is more pronounced for those MLLM-based approaches~\cite{bai2024one,lin2025glus,wang2025object}, which typically require massive multi-source training data to bridge the gap between reasoning and segmentation.
In contrast, our ReferMo demonstrates remarkable data efficiency and inference speed, while achieving significant improvements in long-video understanding.
% We believe it can serve as a promising baseline for future research in long-term RVOS.

% \begin{figure}[t]
%     \centering
%     \includegraphics[width=\linewidth]{fig/motion.pdf} 
%     \caption{Different motion lengths.} 
%     \label{fig:motion} 
% \end{figure}

\begin{table*}[t]
  \centering
  \small
  \caption{Oracle analysis and ablation studies.}\label{tab:ablation}\vspace{-.5em}
  \begin{subtable}{.35\textwidth}
    \centering
    {
    \caption{Oracle analysis with SAM2.} \label{tab:oracle} 
      \setlength{\tabcolsep}{3pt}
      \renewcommand{\arraystretch}{1.25}
      \begin{tabular}{lc|ccc}
        \hline
        \multicolumn{2}{c|}{Dataset} & Point & Box & Mask \\
        \hline
        MeViS~\cite{ding2023mevis} & Valid\_u & 77.3 & 80.0 & 80.6 \\
        \hline
        \multirow{2}{*}{Long-RVOS} & Valid & 54.4 & 55.9 & 56.6 \\
        & Test & 54.3 & 55.6 & 55.6 \\
        \hline
      \end{tabular}
    }
  \end{subtable}\hfill
  \begin{subtable}{.34\textwidth}
    \centering
    {
    \caption{Effect of the video decomposition.} \label{tab:keyframe} 
      \setlength{\tabcolsep}{4pt}
      \renewcommand{\arraystretch}{1.25}%
      \begin{tabular}{l|ccc}
        \hline
        Strategy &$\mathcal{J}$ & $\mathcal{F}$ & $\mathcal{J}\&\mathcal{F}$ \\\hline
        Baseline~\cite{liang2025referdino} & 48.1 & 49.7 & 48.9 \\
        + keyframe. & 49.5 & 50.6 & 50.0 \\
        + keyframe \& motion  & \textbf{50.3} & \textbf{51.8} & \textbf{51.1} \\
        \hline
      \end{tabular}
    }
  \end{subtable}\hfill
  \begin{subtable}{.3\textwidth}
    \centering
    {
      \caption{Different mask propagation methods.}\label{tab:tracker}
      \setlength{\tabcolsep}{4pt}
      \renewcommand{\arraystretch}{1.25}%
      \begin{tabular}{l|ccc}
      \hline
      Method & $\mathcal{J}$ & $\mathcal{F}$ & $\mathcal{J}\&\mathcal{F}$ \\
      \hline
      Xmem++~\cite{bekuzarov2023xmem++} & 49.9 & 51.0 & 50.4 \\
      Cutie~\cite{cheng2024putting} & 49.6 & 50.9 & 50.2 \\
      SAM2~\cite{ravi2024sam2} & \textbf{52.2} & \textbf{53.5} & \textbf{52.9} \\
      \hline
      \end{tabular}
    }
  \end{subtable}\vspace{-1em}
\end{table*}

\vspace{.5em}\noindent\textbf{Fine-grained Evaluation.}
% Long-RVOS provides three types of text descriptions to enable rigorous evaluation.
For most models, the highest performance is achieved on the Static type, followed by Hybrid, and the lowest on Dynamic. This implies a strong bias in current RVOS models toward static attributes, as well as significant challenges in dynamic and temporal understanding.
Furthermore, while $\mathcal{J}\&\mathcal{F}$ scores vary significantly, $\mathrm{tIoU}$ is relatively stable across methods and types. 
This indicates that high $\mathcal{J}\&\mathcal{F}$ scores do not necessarily correlate with strong temporal consistency, and the introduction of $\mathrm{tIoU}$ effectively disentangles these aspects. 
Additionally, the consistently low $\mathrm{vIoU}$ scores across all models suggest that previous evaluations relying solely on frame-averaging metrics may have overestimated the practical robustness of RVOS models.
Against this challenging backdrop, our ReferMo showcases consistent performance improvements over SOTA competitors across all types and metrics.

\vspace{.5em}\noindent\textbf{Oracle Analysis.}
We provide SAM2 with first-frame ground-truth object prompts (i.e., points, boxes or masks) and evaluate its tracking performance.
As shown in \Cref{tab:oracle}, the oracle results for Long-RVOS (54.3$\sim$56.6 $\mathcal{J}\&\mathcal{F}$) are significantly lower than those for MeViS (77.3$\sim$80.6 $\mathcal{J}\&\mathcal{F}$).
The notable performance gap of nearly 25\% demonstrates the long-term challenges in Long-RVOS.

\subsection{Ablation Studies}
\textbf{Effect of Video Decomposition.} 
In contrast to prior RVOS models~\cite{liang2025referdino,yan2024referred,luo2023soc} that directly performs temporal reasoning on the entire video, our ReferMo decomposes the video into clips (keyframe + motion information) to enable local-to-global reasoning. 
To explore the effect of our strategy, we report performance on the keyframes (before SAM2 tracking) in \Cref{tab:keyframe}.
The results show that the keyframe-based decomposition strategy surpasses the baseline by 1.1 $\mathcal{J}\&\mathcal{F}$.
Further incorporating motion information yields an additional +1.1 $\mathcal{J}\&\mathcal{F}$ gain.
Moreover, unlike the baseline, our ReferMo is only trained with keyframe ground truths, yet it achieves much better performance in long-video scenarios. 
These results encourage further exploration of sparse-frame supervision for RVOS task.

% \begin{table}[t]
%     \setlength{\tabcolsep}{5pt}
%     \footnotesize
%     \centering
%     \resizebox{\linewidth}{!}{
%       \begin{tabular}{c|cccccc}
%         \hline
%         Motion Length & 0 & 1 & 3 & 5 & 8 & 11 \\
%         \hline
%         $\mathcal{J}\&\mathcal{F}$ & 49.5 & 49.5 & 49.5 & 49.5 & 49.5 & 49.5 \\
%         \hline
%       \end{tabular}}\vspace{-.5em}
%     \caption{Overall $\mathcal{J}\&\mathcal{F}$ results of different motion lengths.}\label{fig:motion_len}\vspace{-1em}
% \end{table}

% \vspace{.5em}\noindent\textbf{Effect of Motion Information.} We investigate the impact of varying the number of motion frames in ReferMo.
% As shown in \Cref{tab:ablation} (c), the performance without motions is only 49.7 $\mathcal{J}\&\mathcal{F}$.
% However, even using just one motion frame yields +1.6\% $\mathcal{J}\&\mathcal{F}$ improvements.
% Increasing the motion length to 3 frames improves $\mathcal{J}\&\mathcal{F}$ to 51.3, but further increasing only leads to marginal gains.

\begin{table}[t]
\centering
  {
  \setlength{\tabcolsep}{12pt}
  \footnotesize
  \centering
  \caption{Keyframe performance with global interaction.}\label{tab:global}\vspace{-1em}
  \resizebox{\linewidth}{!}{
  \begin{tabular}{cc|cccc}
    \hline
    Global & Motion & $\mathcal{J}$ & $\mathcal{F}$ & $\mathcal{J}\&\mathcal{F}$\\\hline\hline
     &  & 49.8 & 50.4 & 49.8 \\
    \ding{51} & & 49.5 & 50.6 & 50.0 \\
     & \ding{51} & 50.0 & 51.4 & 50.7 \\
    \ding{51} & \ding{51} & 50.3 & 51.8 & 51.1 \\\hline
  \end{tabular}
  }}\vspace{-1em}
\end{table}

\vspace{.5em}\noindent\textbf{Effect of Different Mask Propagation Strategies.} 
We replace SAM2 with other propagation models (e.g., Xmem++~\cite{bekuzarov2023xmem++} and Cuite~\cite{cheng2024putting}) in \Cref{tab:tracker},
which shows that SAM2 contributes 2.5$\sim$2.7 $\mathcal{J}\&\mathcal{F}$ gains to overall performance. 
Notably, by cross-referencing \Cref{tab:test} and \Cref{tab:tracker}, we observe that even combining with these traditional propagation models, our ReferMo still outperforms existing SAM2-based RVOS methods. These results validate the robustness of our approach.

\vspace{.5em}\noindent\textbf{Effects of Global Interaction and Motion Information.} 
In \Cref{tab:global}, we explore the effect of global interaction. 
We observe that a naive local-to-global structure only yields a marginal gain of +0.2 $\mathcal{J}\&\mathcal{F}$ (Row 2 \textit{vs.} Row 1). 
This is because the sparse keyframes provide insufficient context for global reasoning.
In contrast, when we integrate motion features to expand the local window, performance increases significantly by 1.1 $\mathcal{J}\&\mathcal{F}$ (Row 4 \textit{vs.} Row 2).

\begin{table}[t]
    \caption{Overall $\mathcal{J}\&\mathcal{F}$ results at various object occlusion rates.}\label{tab:occlusion}\vspace{-.9em}
    \setlength{\tabcolsep}{3pt}
    \footnotesize
    \centering
    \resizebox{\linewidth}{!}{
      \begin{tabular}{l|cccc}
        \hline
        Method & [0, 0.25] & [0.25, 0.5) & [0.5, 0.75) & [0.75, 1] \\
        \hline\hline
          RGA3~\cite{wang2025object} & 25.6 & 17.8 & 19.1 & 10.4 \\
          MUTR~\cite{yan2024referred} & 47.4 & 38.5 & 30.8 & 17.4 \\
          ReferDINO~\cite{liang2025referdino} & 53.3 & 45.0 & 36.7 & 25.6 \\
          SAMWISE~\cite{cuttano2024samwise} & 39.9 & 39.3 & 38.8 & 38.0 \\\hline
          \textbf{ReferMo} (Ours) & \textbf{54.6} & \textbf{50.6} & \textbf{46.5} & \textbf{39.7} \\
        \hline
      \end{tabular}}\vspace{-1.2em}
\end{table}

\vspace{.5em}\noindent\textbf{Robustness of Keyframe Methods.} 
As a keyframe-based approach, ReferMo may encounter challenges when target objects are absent from selected keyframes.
To evaluate its robustness, we present the results under varying object occlusion rates in \Cref{tab:occlusion}.
The results show that ReferMo consistently outperforms all competitors across all occlusion brackets.
Moreover, as the occlusion rate increases, ReferMo maintains a consistent performance advantage over most methods.
Although its leading margin over SAMWISE~\cite{cuttano2024samwise}, which uses a streaming post-correction mechanism, narrows at high-occlusion scenarios, ReferMo's overall performance is significantly superior.
Therefore, despite relying soly on keyframe reasoning, ReferMo remains sufficiently robust in most non-extreme cases. 

% \subsection{Visualization}
% \section{Limitations and Future Work}

\section{Conclusion}
In this work, we introduce Long-RVOS, a large-scale benchmark for long-term referring video object segmentation, comprising over 2,000 videos averaging 60+ seconds to address the limitations of existing short-term datasets.
To enable comprehensive and rigorous evaluation, we provide three types of descriptions and two novel metrics, $\mathrm{tIoU}$ and $\mathrm{vIoU}$.
Results on Long-RVOS indicate that current RVOS methods struggle severely in long-video scenarios.
Furthermore, we propose ReferMo, a simple motion-enhanced baseline that significantly outperforms existing SOTA methods on long-term videos.
We believe that Long-RVOS and ReferMo will provide a foundation for future research to develop robust models tackling real-world long videos.

%% file: sec/X_suppl.tex
\clearpage
\setcounter{page}{1}

\twocolumn[{
\renewcommand\twocolumn[1][]{#1}% 
\maketitlesupplementary
\begin{center} 
\centering 
\vspace{-1.5em}
\captionsetup{type=table}
\captionof{table}{Definitions of the video attributes.}\label{tab:attr}
\setlength{\tabcolsep}{18pt}
\resizebox{\textwidth}{!}{
    \begin{tabular}{cll}
      \hline
      Attribute & Full Name & Definition \\
      \hline\hline
      \textbf{POC} & Partial Occlusion & The target object is partially occluded in the sequence.\\
      \textbf{FOC} & Full Occlusion & The target object is fully occluded in the sequence.\\
      \textbf{OV} & Out-of-view & The target leaves the video frame completely.\\
      \textbf{LRA} & Long-term Reappearance & Target object reappears after disappearing for at least 100 frames.\\
      \textbf{VC} & View Change &  Viewpoint affects target appearance significantly.\\
      \textbf{ARC} & Aspect Ratio Change & The ratio of bounding box aspect ratio is outside the range [0.5, 2].\\
      \textbf{SV} & Scale Variation & The ratio of any pair of bounding-box is outside of range [0.5,2.0].\\
      \textbf{CM} & Camera Motion & Abrupt motion of the camera.\\
      \textbf{MB} & Motion Blur & The boundary of target object is blurred because of camera or object fast motion.\\
      \hline
    \end{tabular}}
% \vspace{1em}
\captionof{table}{The percentage (\%) of videos featuring specific attributes.}\label{tab:comp_attr}
\setlength{\tabcolsep}{18pt}
  \resizebox{\textwidth}{!}{
  \begin{tabular}{lcccccccccc}
    \hline
    Dataset & POC & FOC & OV & LRA & VC & ARC & SV & CM & MB \\ 
    \hline\hline
    MeViS~\cite{ding2023mevis}     & 54.8 & 15.1 & 28.7 &  0.1 & 10.0 & 88.2 & 78.7 & 49.2 & 18.8 \\
    \textbf{Long-RVOS} (Ours) & \textbf{60.5} & \textbf{36.2} & \textbf{61.0} & \textbf{11.5} & \textbf{25.9} & \textbf{96.2} & \textbf{93.6} & \textbf{60.7} & \textbf{28.7} \\
    \hline
    \end{tabular}}
\end{center}
}]

\section{More Dataset Statistics}
To further highlight the challenges posed by Long-RVOS, we present a statistical analysis of video attributes, with definitions provided in \Cref{tab:attr}.
As shown in \Cref{tab:comp_attr}, compared to the current largest dataset MeViS~\cite{ding2023mevis}, Long-RVOS involves numerous long-video challenges, including frequent object occlusion (POC, FOC, and OV) and long-term object disappearance-reappearance (LRA). In addition, the videos in Long-RVOS exhibit significant object motion (CM and MB) and appearance changes (VC, ARC and SV), making the dataset more akin to real-world scenarios.

\section{More Implementation Details}
\textbf{Motion Extraction.}
Following Video-LaVIT~\cite{jin2024video}, we rely on motion vectors instead of the expensive dense optical flow. 
Formally, given a video clip, we partition each frame into $16\times16$ non-overlapping macroblocks. Then, motion vectors $\vec{m}$ of the $t$-th frame are estimated by matching the macroblocks between the adjacent frames $I_t$ and $I_{t-1}$:
\begin{equation}
  \vec{m}(p,q) = \arg\min_{i,j} \Vert I_t(p,q) - I_{t-1}(p-i,q-j) \Vert,
\end{equation}
where $I(p,q)$ denotes the pixel values of the macroblock at location $(p,q)$, and $(i,j)$ denotes the coordinate offset between the centers of the two macroblocks.
Empirically, the extraction of motion vectors runs at 748 FPS on our device (Intel(R) Xeon(R) Gold 6226R CPU @ 2.90GHz), enabling real-time processing of long videos.

\vspace{1em}\noindent\textbf{Global Interaction.}
This module performs temporal attention over the inter-frame object features, enabling temporal reasoning and understanding.
Since this is a common module in RVOS approaches~\cite{luo2023soc,ding2023mevis,liang2025referdino}, we follow the object-consistent temporal enhancer (OTE) of ReferDINO~\cite{liang2025referdino} rather than designing a new one from scratch.
For clarity, we briefly revist OTE here. 
% Firstly, given $T$-frame object features $\{\mathcal{O}_t\}_{t=1}^T$, OTE utilizes the Hungarian algorithm~\cite{kuhn1955hungarian} to align the objects of each frame with the memory:
% \begin{equation}
% \label{eq:memory_update}
% \hat{\mathcal{O}}_t = \mathrm{Hungarian}\bigl(\mathcal{M}_{t-1},\,\mathcal{O}_t\bigr),
% \quad
% \mathcal{M}_t =
% \begin{cases}
% \mathcal{O}_1, & t = 1,\\[0.5em]
% \bigl(1 - \alpha\,\mathbf{c}^t\bigr)\,\mathcal{M}_{t-1}
% + \alpha\,\mathbf{c}^t\,\hat{\mathcal{O}}_t, & \text{otherwise}.
% \end{cases}
% \end{equation}
% where $\mathcal{M}_t$ denotes the memory at the $t$-th frame, $\alpha$ is the momentum coefficient, $\boldsymbol{c} \in \mathbb{R}^{N_q}$ is the object-sentence relevant scores, and $N_q$ is the number of object queries on each frame.
Given $T$-frame object features $\{O_t\}_{t=1}^T$ where $O_t \in \mathbb{R}^{N_q \times d}$, OTE utilizes the Hungarian algorithm~\cite{kuhn1955hungarian} to align the $N_q$ objects between adjacent frames based on the pairwise cosine similarity.
After that, it performs temporal self-attention over the aligned object features and cross-attention with the sentence features $\widetilde{E}$. We refer the readers to the original paper~\cite{liang2025referdino} for additional details.

\vspace{1em}\noindent\textbf{Training.}
Unlike previous RVOS methods, ReferMo relies only on the keyframe ground-truth annotations for efficient training. Formally, given a text description and a video of $T_c$ clips, ReferMo outputs the prediction sequences $\{\mathbf{p}_i\}_{i=1}^{N_q}$ for the $N_q$ object queries, where each sequence $\boldsymbol{p}_i = \{\hat{\boldsymbol{s}}^t_i, \hat{\boldsymbol{b}}^t_i, \hat{\boldsymbol{m}}^t_i\}_{t=1}^{T_c}$ describes the binary classification probability, bounding box and mask of the $i$-th object query on $t$-th keyframe.
Our training pipeline follows the practice in previous approaches~\cite{wu2022language,luo2023soc,liang2025referdino}.
Suppose $\boldsymbol{y} = \{\boldsymbol{s}^t, \boldsymbol{b}^t, \boldsymbol{m}^t\}_{t=1}^{T_c}$ as the ground truth of keyframes, we select the prediction sequence with the lowest matching cost as the positive and assign the remaining sequences as negative. The matching cost is defined as follows:
\begin{equation}
\begin{split}
    \mathcal{L}_\text{total}\left(\boldsymbol{y}, \boldsymbol{p}_i\right) = & \lambda_{\text{cls}} \mathcal{L}_{\text{cls}}\left(\boldsymbol{y}, \boldsymbol{p}_i \right) + \lambda_{\text{box}} \mathcal{L}_{\text{box}}\left(\boldsymbol{y}, \boldsymbol{p}_i \right) \\
    & + \lambda_{\text{mask}} \mathcal{L}_{\text{mask}}\left(\boldsymbol{y}, \boldsymbol{p}_i \right).
\end{split}
\end{equation}
The matching cost is computed on individual frames and normalized by $T_c$.
Here, $\mathcal{L}_{\text{cls}}$ is the focal loss that supervises the binary classification prediction. $\mathcal{L}_{\text{box}}$ sums up the L1 loss and GIoU loss. $\mathcal{L}_{\text{mask}}$ is the combination of DICE loss, binary mask focal loss and projection loss~\cite{tian2020conditional}.
$\lambda_{\text{cls}}$, $\lambda_{\text{box}}$ and $\lambda_{\text{mask}}$ are scalar weights of individual losses.
The model is optimized end-to-end by minimizing the total loss $\mathcal{L}_\text{total}$ for positive sequences and only the classification loss $\mathcal{L}_{\text{cls}}$ for negative sequences. 

\vspace{1em}\noindent\textbf{Inference.}
ReferMo produces instance mask for the referring object on keyframes and then employs SAM2~\cite{ravi2024sam2} for subsequent frames.   
Specifically, for the prediction sequences $\{\mathbf{p}_i\}_{i=1}^{N_q}$, we select the best sequence with the highest average classification score  as follows:
\begin{equation}
    \sigma=\underset{i \in [1, N_q]}{\arg \max } \frac{1}{T_c} \sum\nolimits_{t=1}^{T_c} \hat{\boldsymbol{s}}^t_i
\end{equation}
Then, the output mask sequence on keyframes is formed as $\{\boldsymbol{m}_\sigma^t \}_{t=1}^{T_c}$. 
For the $t$-th video clip, we use the keyframe prediction $\boldsymbol{m}_\sigma^t$ as the mask prompt for SAM2, which tracks the target across the subsequent frames within the clip.

\section{Validation Results}
In \Cref{tab:valid}, we present the benchmark results on Long-RVOS validation set.
The results show that our ReferMo achieves consistent improvements over previous RVOS methods, especially those built on SAM or SAM2.

\begin{table*}[t]
  \setlength{\tabcolsep}{4pt}
  \footnotesize
  \centering
    \caption{Comparison on Long-RVOS valid set. FPS is estimated at 360P on Nvidia A6000 GPUs, excluding the video loading time.}\label{tab:valid}\vspace{-.5em}
  \resizebox{\textwidth}{!}{
    \begin{tabular}{l |ccc |ccc |ccc |ccc |c}
      \hline
      \multirow{2}{*}{Method} & \multicolumn{3}{c|}{Static} & \multicolumn{3}{c|}{Dynamic} & \multicolumn{3}{c|}{Hybrid} & \multicolumn{3}{c|}{Overall} & \multirow{2}{*}{FPS} \\ 
      & $\mathcal{J}\&\mathcal{F}$ & $\mathrm{tIoU}$ & $\mathrm{vIoU}$ & $\mathcal{J}\&\mathcal{F}$ & $\mathrm{tIoU}$ & $\mathrm{vIoU}$ & $\mathcal{J}\&\mathcal{F}$ & $\mathrm{tIoU}$ & $\mathrm{vIoU}$ & $\mathcal{J}\&\mathcal{F}$ & $\mathrm{tIoU}$ & $\mathrm{vIoU}$ & \\
      \hline\hline
      {\scriptsize\color{cvprblue}\textit{Without SAM / SAM2}} &&&&&&&&&&&& \\
      SOC~\cite{luo2023soc} {\scriptsize\color{gray}NeurIPS'23} & 38.7 & 73.1 & 34.9 & 37.8 & 74.6 & 34.1 & 37.8 & 74.3 & 34.5 & 38.1 & 74.0 & 34.5 & 53.8 \\
      MUTR~\cite{yan2024referred} {\scriptsize\color{gray}AAAI'24} & 44.1 & 73.5 & 40.3 & 42.0 & 75.2 & 38.9 & 43.5 & 74.6 & 40.2 & 43.2 & 74.4 & 39.8 & 20.4 \\
      ReferDINO~\cite{liang2025referdino} {\scriptsize\color{gray}ICCV'25} & 52.5 & \textbf{74.2} & 48.2 & 46.7 & \textbf{75.2} & 42.9 & 49.3 & \textbf{74.8} & 45.4 & 49.6 & \textbf{74.7} & 45.6 & 46.4 \\
      \hline
      {\scriptsize\color{cvprblue}\textit{With SAM / SAM2}} &&&&&&&&&&&& \\
      VideoLISA~\cite{bai2024one} {\scriptsize\color{gray}NeurIPS'24} & 17.3 & 66.8 & 12.7 & 12.9 & 72.6 & 6.8 & 12.1 & 72.3 & 6.0 & 14.1 & 70.5 & 8.6 & 6.6 \\
      GLUS~\cite{lin2025glus} {\scriptsize\color{gray}CVPR'25} & 24.4 & 62.8 & 20.8 & 26.1 & 64.7 & 23.1 & 24.1 & 63.5 & 20.6 & 24.8 & 63.7 & 21.5 & 3.6 \\
      SAMWISE~\cite{cuttano2024samwise} {\scriptsize\color{gray}CVPR'25} & 42.3 & 61.5 & 31.2 & 40.7 & 63.3 & 31.4 & 40.6 & 65.8 & 31.2 & 41.2 & 63.5 & 31.2 & 7.0 \\
      RGA3~\cite{wang2025object} {\scriptsize\color{gray}ICCV'25} & 21.1 & 61.0 & 15.4 & 22.3 & 62.8 & 17.5 & 21.1 & 61.8 & 16.4 & 21.5 & 61.8 & 16.4 & 8.7 \\
      \hline
      \rowcolor{cvprblue!10} \textbf{ReferMo} (Ours) & \textbf{56.7} & 74.0 & \textbf{49.4} & \textbf{50.7} & 74.2 & \textbf{43.4} & \textbf{53.7} & 74.7 & \textbf{47.4} & \textbf{53.7} & 74.3 & \textbf{46.8} & 52.5 \\
      \hline
    \end{tabular}}
\end{table*}

% \begin{table}[t]
% \centering
%   {
%   \setlength{\tabcolsep}{14pt}
%   \footnotesize
%   \centering
%   \resizebox{\linewidth}{!}{
%   \begin{tabular}{cc|cccc}
%     \hline
%     Global & Motion & $\mathcal{J}$ & $\mathcal{F}$ & $\mathcal{J}\&\mathcal{F}$\\\hline\hline
%      &  & 49.8 & 50.4 & 49.8 \\
%     \ding{51} & & 49.5 & 50.6 & 50.0 \\
%      & \ding{51} & 50.0 & 51.4 & 50.7 \\
%     \ding{51} & \ding{51} & 50.3 & 51.8 & 51.1 \\\hline
%   \end{tabular}
%   }}
%   \caption{Keyframe performance with global interaction.}\label{tab:global}
% \end{table}

% \begin{table}
%   \captionof{table}{Keyframe performance with different global interaction methods. \textit{None} indicates that the global interaction is removed.}
%   \label{tab:global}
%   \fontsize{8}{10}\selectfont
%   \setlength{\tabcolsep}{12pt}
%   \resizebox{\linewidth}{!}{
%       \begin{tabular}{c|ccc}
%         \toprule
%         Method & $\mathcal{J}\&\mathcal{F}$ & $\mathcal{J}$ & $\mathcal{F}$ \\\midrule
%         \textit{None} & 47.2 & 45.6 & 48.7 \\\midrule
%         LMPM~\cite{ding2023mevis} & 48.9 & 47.3 & 50.5\\
%         OTE~\cite{liang2025referdino} & 49.6 & 48.0 & 51.2 \\\bottomrule
%   \end{tabular}}
% \end{table}

\section{More Ablation Studies}
\textbf{Effectiveness of Gating Image-Motion Fusion.} 
ReferMo employs the spatial-aware gating (SG)  and channel-aware gating (CG) mechanisms in image-motion fusion to avoid undesired motion noise. 
As shown in \Cref{tab:fusion}, directly concatenating keyframe and motion features leads to a performance collapse. 
This is because RVOS requires per-frame fine-grain perception, while directly integrating motion features can introduce significant object-irrelevant noise.
By applying these two gating strategies, ReferMo effectively alleviates such noise while preserving only the motion cues that highlight target objects, thereby yielding significant performance gains.

% \vspace{1em}\noindent\textbf{Effects of Global Interaction and Motion Information.} 
% In \Cref{tab:global}, we explore the effect of global interaction. 
% We observe that a naive local-to-global structure only yields a marginal gain of +0.2 $\mathcal{J}\&\mathcal{F}$ (Row 2 \textit{vs.} Row 1). 
% This is because the sparse keyframes provide insufficient context for global reasoning.
% In contrast, when we integrate motion features to expand the local window, performance increases significantly by 1.1 $\mathcal{J}\&\mathcal{F}$ (Row 4 \textit{vs.} Row 2).

\begin{table}[t]
\centering
  \caption{Keyframe results of different image-motion fusion approaches.}\label{tab:fusion}\vspace{-.5em}
  {
  \setlength{\tabcolsep}{10pt}
  \footnotesize
  \centering
  \resizebox{\linewidth}{!}{
  \begin{tabular}{cc|ccc}
    \hline
    SG & CG & $\mathcal{J}$ & $\mathcal{F}$ & $\mathcal{J}\&\mathcal{F}$\\\hline\hline
     & & 28.8 & 28.7 & 28.7 \\\hline
    \ding{51} & & 46.9 & 46.3 & 46.6\\
     & \ding{51} & 49.5 & 50.4 & 50.0\\
    \ding{51} & \ding{51} & 50.3 & 51.8 & 51.1 \\\hline
  \end{tabular}
  }}
\end{table}

\begin{table}[t]
\centering
  \caption{Overall $\mathcal{J}\&\mathcal{F}$ results for different description lengths.}\label{tab:length}\vspace{-.5em}
  {
  \setlength{\tabcolsep}{10pt}
  \footnotesize
  \centering
  \resizebox{\linewidth}{!}{
  \begin{tabular}{l|ccc}
    \hline
    Method & \textbf{$<$10} & \textbf{[10, 20]} & \textbf{$>$20} \\ \hline\hline
    RGA3~\cite{wang2025object} & 23.8 & 22.6 & 19.8 \\
    MUTR~\cite{yan2024referred} & 42.5 & 43.1 & 38.2 \\
    SAMWISE~\cite{cuttano2024samwise} & 40.1 & 41.5 & 40.8 \\
    ReferDINO~\cite{liang2025referdino} & 49.2 & 49.2 & 44.3 \\\hline
    ReferMo (ours) & \textbf{53.6} & \textbf{53.6} & \textbf{48.5} \\
    \hline
  \end{tabular}
  }}
\end{table}

\vspace{1em}\noindent\textbf{Effect of Description Length.} 
We evaluate the impact of varying description lengths and present the results in Table R1. As description length increases, slight performance declines are observed across models. However, our ReferMo consistently outperforms existing methods across different description lengths.

\begin{table}[t]
\centering
  {
  \setlength{\tabcolsep}{2pt}
  \footnotesize
  \centering
  \caption{Overall $\mathcal{J}\&\mathcal{F}$ results by event complexity.}\label{tab:event}\vspace{-.5em}
  \resizebox{\linewidth}{!}{
  \begin{tabular}{l|ccc}
    \hline
    Method & \textbf{\textit{Single-event}} & \textbf{\textit{Two-event}} & \textbf{\textit{Multi-event}} \\ \hline\hline
    RGA3~\cite{wang2025object} & 23.0 & 22.6 & 19.2 \\
    MUTR~\cite{yan2024referred} & 42.7 & 40.0 & 36.0 \\
    SAMWISE~\cite{cuttano2024samwise} & 40.6 & 41.7 & 38.0 \\
    ReferDINO~\cite{liang2025referdino} & 48.4 & 44.7 & 37.2 \\\hline
    ReferMo (ours) & \textbf{52.9} & \textbf{48.0} & \textbf{40.5} \\
    \hline
  \end{tabular}
  }}
\end{table}

\vspace{1em}\noindent\textbf{Effect of Multi-event Videos.}
To explore the impact of event number in a video on model performance, we categorized the samples into single-event, two-event, and multi-event groups based on the keywords (e.g., \textit{then}, \textit{finally}, \textit{ultimately}) in descriptions. As shown in \Cref{tab:event}, performance across models declines as the event number increases, yet our ReferMo consistently outperforms existing methods. Also, these results highlight the significance of our long-term benchmark for evaluating the capabilities of RVOS models in understanding complex event sequences.

\begin{figure*}[t]
    \centering
    \includegraphics[width=\textwidth]{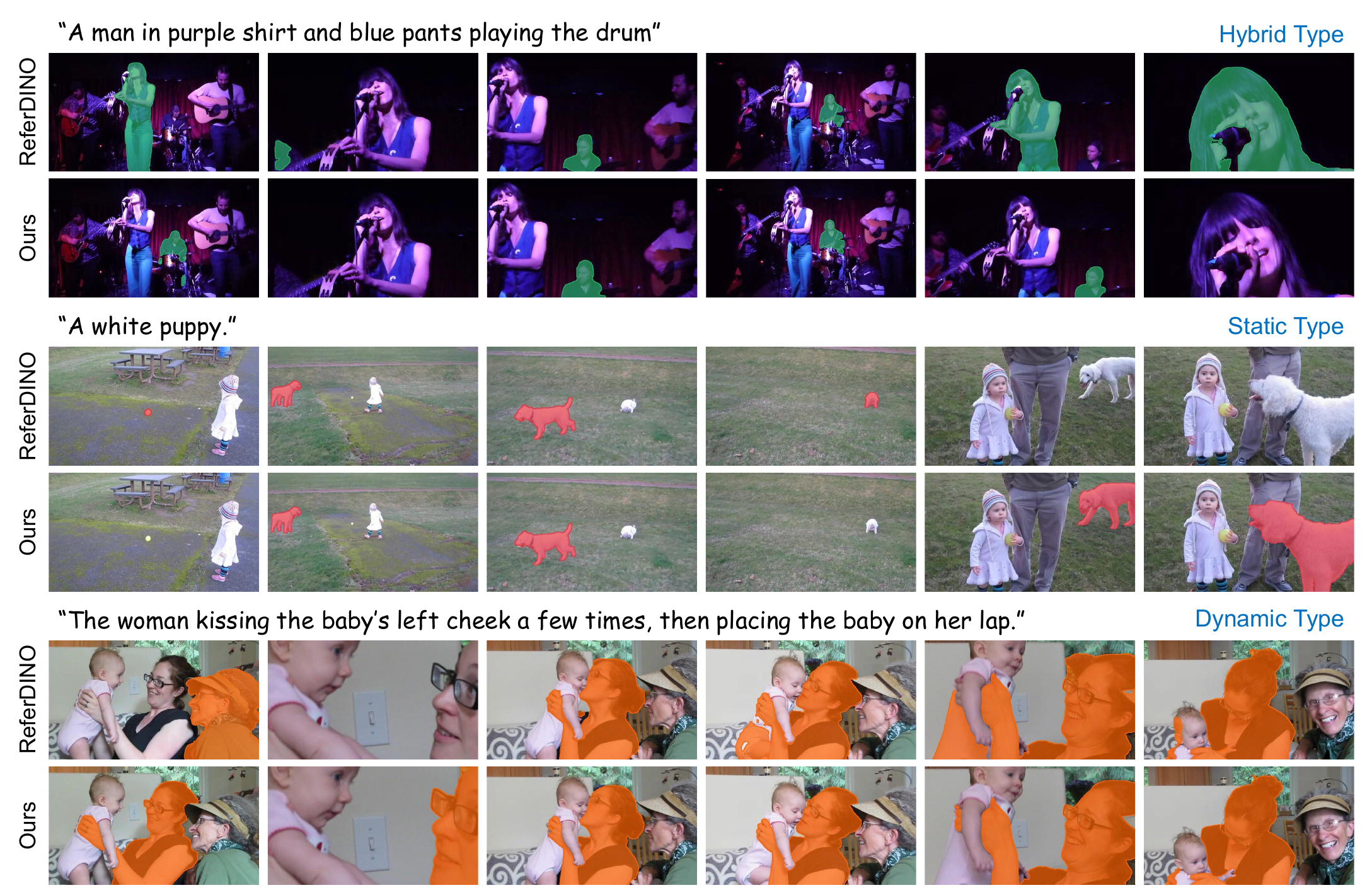}
    \caption{Qualitative comparison of our ReferMo with the SOTA model ReferDINO~\cite{liang2025referdino}. ReferMo performs better in long-term object consistency and segmentation quality.
    }
    \label{fig:visual}
\end{figure*}

\section{Visualization}
In \Cref{fig:visual}, we provide the qualitative comparisons with the SOTA model ReferDINO~\cite{liang2025referdino} on Long-RVOS. These examples involve multiple long-term challenges, such as object occlusion, disappearance-reappearance and view changes. 
The results clearly show the effectiveness of our baseline ReferMo in long-term object consistency and segmentation quality. 

\section{Limitations and Future Work}
In this work, we chose to begin with description-based RVOS because it is commonly used in current video applications and this task remains far from being solved.
It is promising to broaden the benchmark scope to support more tasks, such as reasoning RVOS~\cite{bai2024one,yan2024visa}, semi-supervised VOS~\cite{pont20172017,xu2018youtube,ding2023mose}, interactive VOS~\cite{kirillov2023segment,ravi2024sam2} and audio-guided VOS~\cite{yan2024referred}. 
Besides, while our benchmark covers a variety of objects, it currently does not include background stuff classes (e.g., sky and river), which could be incorporated in future work for covering more diverse scenarios.
% For training efficiency, our simple baseline only employs SAM2 at inference.
% Future work may explore parameter-efficient fine-tuning techniques to enable end-to-end optimization with SAM2.

% \section{Border Impacts}
% In this work, we propose a large scale benchmark to advance the RVOS task toward long-term, real-world scenarios.
% This benchmark potentially leads to the development of stronger and more practical RVOS models. 
% These models hold significant potential for many real-world applications, such as video editing, human-robot interaction and automated video analysis.
% However, like many powerful AI technologies, RVOS models carry potential risks, including unauthorized surveillance or privacy-infringing tracking.
% Despite these concerns, we firmly believe that the task itself is neutral, and its positive implications outweigh potential risks when guided by ethical considerations and responsible deployment.